\documentclass[conference]{IEEEtran}
\IEEEoverridecommandlockouts
\usepackage{cite}
\usepackage{amsmath,amssymb,amsfonts}
\usepackage{algorithmic}
\usepackage{graphicx}
\usepackage{textcomp}
\usepackage{xcolor}

\usepackage{bm}
\usepackage{amsthm}
\usepackage{mathrsfs}

\makeatletter
\theoremstyle{plain}

\theoremstyle{remark}

\theoremstyle{plain}

\theoremstyle{plain}
\newtheorem{prop}{\protect\propositionname}

\makeatother
\usepackage[english]{babel}
\providecommand{\lemmaname}{Lemma}
\providecommand{\remarkname}{Remark}
\providecommand{\theoremname}{Theorem}
\providecommand{\propositionname}{Proposition}

\newcommand{\R}{\mathbb{R}}

\newcommand{\fnorm}[1]{\left\lVert#1\right\rVert_F}

\newcommand{\lowert}[1]{\left\lfloor#1\right\rfloor}

\newcommand{\matlog}{\textbf{\text{Log}}}
\newcommand{\matexp}{\textbf{\text{Exp}}}
\newcommand{\matdiag}{\textbf{\text{D}}}
\newcommand{\cholspace}[1]{\mathbb{L}_{#1}}

\newcommand{\manifold}{\mathcal{M}}
\newcommand{\chol}{\mathscr{L}}
\newcommand{\spd}[1]{\mathbb{S}_{#1}}
\newcommand{\metric}{g}
\newcommand{\prob}{P}
\newcommand{\dist}{d}

\newcommand{\set}[1]{\left\{#1\right\}}
\newcommand{\sample}{\mathcal{Z}}
\newcommand{\spdsample}{\mathcal{Z}}

\newcommand{\E}{\mathbb{E}}
\newcommand{\rv}{Q}

\DeclareMathOperator{\vvec}{\text{vec}}

\def\BibTeX{{\rm B\kern-.05em{\sc i\kern-.025em b}\kern-.08em
    T\kern-.1667em\lower.7ex\hbox{E}\kern-.125emX}}
\begin{document}

\title{
%$k$-means on the log-Cholesky manifold with Symmetric Positive Definite matrices\\
%$k$-means on a log-Cholesky Manifold, with Unsupervised Classification of Radar Products \\
$k$-means on Positive Definite Matrices, and an Application to Clustering in Radar Image Sequences 
}

% \author{\IEEEauthorblockN{1\textsuperscript{st} Daniel Fryer}
% \IEEEauthorblockA{\textit{School of Mathematics and Physics} \\
% \textit{The University of Queensland.}\\
% St Lucia, Australia \\
% ORCID 0000-0001-6032-0522}
% \and
% \IEEEauthorblockN{2\textsuperscript{nd} Hien Nguyen}
% \IEEEauthorblockA{\textit{Department of Mathematics and Statistics} \\
% \textit{La Trobe University.}\\
% Bundoora, Australia \\
% ORCID 0000-0002-9958-432X}
% \and
% \IEEEauthorblockN{3\textsuperscript{rd} Pascal Castellazzi}
% \IEEEauthorblockA{\textit{Land and Water Deep Earth Imaging FSP} \\
% \textit{Commonwealth Scientific and Industrial Research Organisation}\\
% Urrbrae, SA, Australia \\
% ORCID: 0000-0002-5591-0867}
%
%%\and
%%\IEEEauthorblockN{4\textsuperscript{th} Given Name Surname}
%%\IEEEauthorblockA{\textit{dept. name of organization (of Aff.)} \\
%%\textit{name of organization (of Aff.)}\\
%%City, Country \\
%%email address or ORCID}
%%\and
%%\IEEEauthorblockN{5\textsuperscript{th} Given Name Surname}
%%\IEEEauthorblockA{\textit{dept. name of organization (of Aff.)} \\
%%\textit{name of organization (of Aff.)}\\
%%City, Country \\
%%email address or ORCID}
%%\and
%%\IEEEauthorblockN{6\textsuperscript{th} Given Name Surname}
%%\IEEEauthorblockA{\textit{dept. name of organization (of Aff.)} \\
%%\textit{name of organization (of Aff.)}\\
%%City, Country \\
%%email address or ORCID}
% }

\author{
\IEEEauthorblockN{Daniel Fryer\IEEEauthorrefmark{1},
Hien Nguyen\IEEEauthorrefmark{2} and
Pascal Castellazzi\IEEEauthorrefmark{3}}
\IEEEauthorblockA{\IEEEauthorrefmark{1}School of Mathematics and Physics\\
The University of Queensland, St Lucia, Australia\\
ORCID 0000-0001-6032-0522}
\IEEEauthorblockA{\IEEEauthorrefmark{2}Department of Mathematics and Statistics\\
La Trobe University, Bundoora, Australia\\
ORCID 0000-0002-9958-432X}
\IEEEauthorblockA{\IEEEauthorrefmark{3}Deep Earth Imaging FSP, Land and Water\\
Commonwealth Scientific and Industrial Research Organisation (CSIRO), Urrbrae, SA, Australia\\
ORCID: 0000-0002-5591-0867}}

\maketitle

%-------------------------------------------------------------------------------
\begin{abstract}
We state theoretical properties for $k$-means clustering of Symmetric Positive Definite (SPD) matrices, in a non-Euclidean space, that provides a natural and favourable representation of these data. We then provide a novel application for this method, to time-series clustering of pixels in a sequence of Synthetic Aperture Radar images, via their finite-lag autocovariance matrices.
\end{abstract}


%-------------------------------------------------------------- sec:introduction
\section{Introduction} \label{sec:introduction}

Many objects of interest in applied mathematics and engineering can be represented, often uniquely, by a Symmetric Positive Definite (SPD) matrix. For example, SPD matrices correspond bijectively to mean centered Gaussian distributions, and are used to model Brownian motion in Diffusion Tensor Imaging (DTI), where they are referred to as tensors \cite{[18]}. The finite-lag autocovariance matrices of time-series are SPD, and have been used in compression based clustering \cite{aghabozorgi2015time}, for analysing dynamical brain functional connectivity \cite{dai2019analyzing}, and in our application (Section \ref{sec:application}). Many more examples are mentioned in \cite{[5],[18]}.

For a given $m$, the space of $m \times m$ SPD matrices forms the interior of a blunt convex cone in $\R^{m(m+1)/2}$, and is not a vector space under addition and scalar multiplication. Thus, many standard algorithms applied to SPD matrices, with the Euclidean norm, may produce symmetric matrices that are not positive definite, having non-positive eigenvalues \cite{[18]}. Efforts to avoid this shortcoming have led to multiple suggestions for alternative Riemannian metrics, that may endow the space with a more favourable structure. The most popular of these to date has been the affine-invariant metric \cite{[20]}, also known in statistics as the Fisher-Rao metric \cite{calvo1991explicit}. Unfortunately, there is no closed form for the Fr\'echet mean under this metric, though an MM algorithm implementation is given in \cite{[7]}. The log-Euclidean metric \cite{[3],[11]} arose from attempts to put a Lie group structure on SPD matrices, thus transferring across the vector space structure of symmetric matrices, while preserving many affine-invariant qualities \cite{[18]}. Recently, in \cite{[5]}, a new metric was introduced: the log-Cholesky metric, based on the Cholesky decomposition, a diffeomorphism between upper triangular positive definite and SPD matrices. In this work, we focus on $k$-means clustering of SPD matrices on this metric space. 

Clustering via $k$-means can be used to extract information regarding heterogeneity of matrix variate data in a computationally efficient manner. Standard Euclidean $k$-means clustering, minimising within-cluster variation, is uncomplicated, gives asymptotically normal \cite{[12]} and strongly consistent \cite{[6]} cluster centers, and can be scaled to massive and distributed data \cite{jin2006fast}. However, performing $k$-means on SPD matrices may be difficult, without a computationally efficient form for the Fr\'echet mean \cite{[15]}.

In Section \ref{sec:logchol}, we introduce the log-Cholesky distance and closed-form expression for the corresponding Fr\'echet mean. We then identify a diffeomorphism under which the log-Cholesky metric norm reduces to the Euclidean norm in $\R^{m(m+1)/2}$.  We then use this fact to prove that $k$-means on the log-Cholesky manifold satisfies the same consistency and asymptotic normality properties as Euclidean $k$-means. Also, we show that the average objective function converges towards its optimal value, almost surely, at a rate of $(\log(n)/n)^{1/2}$. 

In Section \ref{sec:application}, in a demonstrative application of these results, we perform $k$-means time-series clustering via finite lag autocovariance matrices, representing pixels in a sequence of Synthetic Aperture Radar (SAR) images of the Mount Gambier region of South Australia. This is done efficiently by leveraging existing low-level software libraries for computing Euclidean $k$-means, over a sample of $2{,}929{,}052$ time-series, with multiple passes for parameter tuning. This results in suggestions for improvement on previous work of \cite{castellazzi2019towards}, predicting the locations of Groundwater Dependent Vegetation (GDV). A brief discussion is provided in Section \ref{sec:discussion}.

%---------------------------------------------------------------- subsec:def-kms
\subsection{Definition of $k$-means and Fr\'echet mean} \label{subsec:def-kms}

A manifold is a set of points that is everywhere locally homeomorphic (or, loosely speaking, is smoothly deformable) to a subset of a Euclidean space. A Riemannian manifold is a manifold $\manifold$, that is equipped with a globally defined differential structure, to allow calculus to be performed, and a Riemannian metric $g$, so that angle and length can be defined \cite{[18]}. Given a random element $\rv$ with distribution $\prob$ on a Riemannian manifold $(\manifold,\metric)$, with distance function $\dist$, the classical generalisation of the Euclidean centre of mass \cite{[3], [9]} is to define the set of Karcher means,
\begin{equation} \nonumber 
\set{k \, :\, k =   \arg\min_{x \in \manifold}\E \, \dist^2(a,\rv)} \subseteq \manifold,
\end{equation}
as the set of points in the manifold that minimise the dispersion
\[
\E\, \dist^2(a,\rv) = \int_\manifold \!\! \dist^2(a,q) \, d\prob(q).
\]
When a unique minimiser exists, it is called the Fr\'echet mean $\E Q$. If the distribution $\prob$ is sufficiently localised, then the existence of $\E Q$ is guaranteed \cite{[8]}. Given a finite set $\sample \subseteq \manifold$ of points on the manifold, we can define the empirical Fr\'echet mean
\begin{equation} \label{eq:frechet}
\overline{S} = \arg\min_{a \in \manifold} \sigma^2(a, \sample),
\end{equation}
as the minimiser of the empirical dispersion
\[
\sigma^2(a, \sample) = \sum_{S \in \sample} d^2(a,S).
\]
In $k$-means clustering, we seek to find a partition of $\sample$ into disjoint subsets $\mathcal{K}=\{\mathcal{Z}_1,\ldots,\mathcal{Z}_k\} \subseteq \sample$ (some of which may be empty), minimising the overall sum of squared distances
\begin{equation} \label{eq:D}
\mathcal{D}^k(\mathcal{K}) = \sum_{j=1}^k \sigma^2(\overline{S}_j, \mathcal{Z}_j),
\end{equation}
where $\overline{S}_j$ is the empirical Fr\'echet mean of the cluster $\mathcal{Z}_j \subseteq \mathcal{Z}$. The $k$-means objective can be reinterpreted in terms of finding the \textit{centroids} $\overline{S}_j$. That is, we search for $k$ (possibly non-distinct) centroids minimising the nearest neighbour dispersion \eqref{eq:D}, where $\mathcal{Z}_j$ contains the points with nearest centroid $\overline{S}_j$. Both phrasings -- finding the centroids or finding the partition -- are equivalent, since the Fr\'echet mean is the dispersion minimising centroid \eqref{eq:frechet}. 

%---------------------------------------------------------------- subsec:def-smd
\subsection{Symmetric Positive Definite Matrices} \label{subsec:def-smd}

A Symmetric Positive Definite (SPD) matrix $S$ is a square symmetric matrix with real entries satisfying ${x^TSx > 0}$ for all vectors in $x \in \R^m \setminus \{\bm{0}\},$ where $S$ is $m \times m.$ Equivalently, a square symmetric matrix is positive definite if all of its eigenvalues are positive. SPD matrices can be understood geometrically as encoding ellipsoids, or scaling along a set of $m$ orthonormal basis vectors in $\R^m$. That is, $S$ can be decomposed as $S = U^{T}DU,$ where $U$ is orthogonal and $D$ is diagonal. In particular, this implies a simple expression \cite{[5]} for calculating an arbitrary analytic matrix function $\mathbf{f}$, such as the matrix logarithm $\matlog$, or exponential $\matexp$, as $\mathbf{f}(S) = U^{T}\mathbf{f}(D)U$ where $\mathbf{f}(D)$ is the diagonal matrix with $i$th diagonal entry $\mathbf{f}(D)_{ii} = \mathbf{f}(D_{ii}).$ 

The space $\spd{m}$ of SPD matrices is closed under addition and multiplication by positive real numbers, but not under multiplication by non-negative real numbers, thus forming the interior of a blunt convex cone \cite{[18]}. However, every $S \in \spd{m}$ permits a Cholesky decomposition $S = LL^T$, where $L$ is a lower triangular matrix with positive real diagonals \cite{[5]}, and in Section \ref{subsec:def-logchol} we use the Cholesky decomposition to construct a diffeomorphism between $\spd{m}$ and $\R^{m(m+1)/2}.$

% %---------------------------------------------------------------- subsec:def-sar
% \subsection{Synthetic Aperture Radar} \label{subsec:def-sar}

%------------------------------------------------------------------- sec:logchol
\section{Log-Cholesky $k$-means} \label{sec:logchol}

%------------------------------------------------------------ subsec:def-logchol
\subsection{Log-Cholesky distance and mean} \label{subsec:def-logchol}
We use $\cholspace{m}$ to denote the space of lower triangular matrices with positive diagonal. The map $\chol : \spd{m} \rightarrow \cholspace{m}$, that sends an SPD matrix to its Cholesky factor, was shown in \cite[Proposition 2]{[5]} to be a diffeomorphism. So, with this one-to-one correspondence in mind, define for $L,K \in \cholspace{m}$ the distance function, 
\begin{align} \label{eq:dc}
d_C^2(L,K) & = \fnorm{\lowert{L} - \lowert{K}}^2 \nonumber \\
 & + \fnorm{\matlog(\matdiag(L)) - \matlog(\matdiag(K))}^2, 
\end{align}
where $\fnorm{\cdot}$ is the Frobenius (i.e., vectorised Euclidean) norm and $\matdiag$ is the diagonalisation function that maps off diagonal elements to $0$. In \cite[Proposition 10]{[5]} it was also shown that, under this distance function, the Fr\'echet mean of a random SPD matrix $S$ exists, is unique, and takes the following closed form, provided that $\E d^2_C(L,\chol S) < \infty$ for some $L \in \cholspace{m}$.
\begin{equation}
\E S = \chol^{-1}\!\Big[\,\, 
\E\!\lowert{\chol S} + 
\matexp\!\set{\,
\E\matlog(\matdiag(\chol S )
\,}\,\,\Big].    
\end{equation}
It follows \cite[Corollary 12]{[5]} that a subset $\spdsample \subseteq \spd{m}$ has empirical Fr\'echet mean $\overline{S}$ given by inverting
\begin{align} \label{eq:lcmean}
\chol(\overline{S}) & =\sum_{S \in \spdsample} \frac{\lowert{\chol S}}{|\spdsample|} +  \matexp\set{\sum_{S \in \spdsample} \frac{\matlog(\matdiag(\chol S)) }{|\spdsample|}} .
\end{align}

%----------------------------------------------------------------- subsec:euclid
\subsection{Reduction to Euclidean mean} \label{subsec:euclid}
Define the map $\mathcal{V} : \spd{m} \rightarrow \R^{m(m+1)/2}$ given by
\begin{equation} \label{eq:log-chol-vec}
  \mathcal{V}(S) = \left( \ell_1, \ldots, \ell_{m(m-1)/2}, d_1,\ldots,d_m \right)^T,
\end{equation}
where $d_i = \log(\chol S_{ii})$, the log transformed $i$th diagonal element of $\chol S,$ and $\ell_i = (\vvec \lowert{\chol S})_i$, the $i$th coordinate output of the vectorisation operator, applied to the lower triangle of $\chol S$. So, $\mathcal{V}$ is a composition of the Cholesky map $\chol$, the scalar logarithm, and a vectorisation that drops the (vanishing) upper triangle elements. Hence, $\mathcal{V}$ is bijective and continuous, since $\chol$ is a diffeomorphism from $\spd{m}$ to $\cholspace{m}$. It follows that $\mathcal{V}$ preserves all compact sets. Furthermore, we now have that \eqref{eq:dc} reduces to the Euclidean distance.
So, \eqref{eq:lcmean} can be written,
\begin{equation}
\overline{S} 
= 
\mathcal{V}^{-1}\left(\frac{1}{|\mathcal{Z}|} \sum_{S \in \mathcal{Z}} \mathcal{V}(S)
\right).
\end{equation}
In other words, $\mathcal{V}(S)$ provides a one-to-one continuous mapping between the space of SPD matrices and a Euclidean space, where the Log-Cholesky Fr\'echet mean reduces to the Euclidean mean. It follows that the corresponding Fr\'echet $k$-means is exactly the same as a Euclidean $k$-means. 

%----------------------------------------------------------------- subsec:theory
\subsection{Theoretical results} \label{subsec:theory}
For fixed $k$, we can write the $k\text{-means}$ objective function
for $n$ observations $S_{i}$, $i\in\left[n\right]$, as
\[
\mathcal{D}_{n}^{k}\left(X^{k}\right)=\frac{1}{n}\sum_{i=1}^{n}\min_{j\in\left[k\right]}\left\Vert \mathcal{V}\left(S_{i}\right)-X_{j}\right\Vert ^{2}\text{,}
\]
where $X^{k}=\left\{ X_{1},\dots,X_{k}\right\} $, and $X_{j}\in\mathbb{R}^{m\left(m+1\right)/2}$,
for each $j\in\left[m\right]$. Let
\[
X_{n}^{k}=\left\{ X_{n,1}^{k},\dots,X_{n,k}^{k}\right\} =\underset{X}{\arg\min}\,\mathcal{D}_{n}^{k}\left(X^{k}\right)\text{;}
\]
then, with some abuse of notation, the set of optimal cluster centers
in $\mathbb{S}_{m}$ is:
\[
\mathcal{V}^{-1}\left(X_{n}^{k}\right)=\left\{ \mathcal{V}^{-1}\left(X_{n,1}^{k}\right),\dots,\mathcal{V}^{-1}\left(X_{n,k}^{k}\right)\right\} \text{.}
\]
Let $P$ be a probability measure on the set $\mathbb{S}_{m}$, and
let $A$ be a finite subset of $\mathbb{R}^{m\left(m+1\right)/2}$.
Further, define
\[
\mathcal{Q}\left(A,P\right)=\mathbb{E}_{P_{S}}\left[\min_{\alpha\in A}\,\left\Vert \mathcal{V}\left(S_{i}\right)-\alpha\right\Vert ^{2}\right]
\]
and $m_{k}\left(P\right)=\inf\left\{ \mathcal{Q}\left(A,P\right):\#\left(A\right)\le k\right\} $.
The following consistency theorem can be obtained via the main theorem
of \cite{[10]}.
\begin{prop}
Assume that $S_{1},\dots,S_{n}$ are IID and arise from a data generating
process with probability measure $P_{S}$, with $\mathbb{E}_{P_{S}}\left\Vert \mathcal{V}\left(S\right)\right\Vert ^{2}<\infty$,
and that for each $j\in\left[k\right]$, there exists a unique set
$A^{j}$, such that $\mathcal{Q}\left(A^{j},P_{S}\right)=m_{j}\left(P_{S}\right)$.
Then, $X_{n}^{k}\rightarrow A^{k}$ and $\mathcal{D}_{n}^{k}\left(X^{k}\right)\rightarrow m_{k}\left(P_{S}\right)$,
almost surely.
\end{prop}
Let $\underline{X}_{n}^{k}$ and $\underline{A}^{k}$ be vectors containing
the elements of $X_{n}^{k}$ and $A^{k}$, respectively, and let $\underline{\mathcal{Q}}\left(\underline{A}^{k},P\right)$
be a vector-input version of $\mathcal{Q}\left(A^{k},P\right)$. Further,
denote the Hessian of $\underline{\mathcal{Q}}$, with respect to
$\underline{A}^{k}$, by $\mathbb{H}\left[\underline{\mathcal{Q}}\left(\underline{A}^{k},P\right)\right]$.
We can deduce the asymptotic normality result regarding $\underline{X}_{n}^{k}$
via the main theorem of \cite{[12]}.
\begin{prop}
In addition to the conditions of Proposition 1, assume that $P_{S}$
can be characterized by a probability density function $f_{\mathcal{V}}$,
with respect to the transformation $\mathcal{V}\left(S\right)\in\mathbb{R}^{m\left(m+1\right)/2}$,
where $f_{\mathcal{V}}\left(S\right)\le h\left(\left\Vert \mathcal{V}\left(S\right)\right\Vert \right)$,
for all $S\in\mathbb{S}_{m}$, such that $\int_{0}^{\infty}r^{m\left(m+1\right)/2}h\left(r\right)$dr, for some dominating function $h$.
If we further assume that $\mathbb{H}\left[\underline{\mathcal{Q}}\left(\underline{A}^{k},P_{S}\right)\right]$
is positive definite, then $n^{-1/2}\left(\underline{X}_{n}^{k}-\underline{A}_{k}\right)$
is asymptotically normal with mean $0$ and covariance $\left[\mathbb{H}\left[\underline{\mathcal{Q}}\left(\underline{A}^{k},P_{S}\right)\right]\right]^{-1}\Sigma\left[\mathbb{H}\left[\underline{\mathcal{Q}}\left(\underline{A}^{k},P_{S}\right)\right]\right]^{-1}$,
where $\Sigma$ is a $km\left(m+1\right)/2\times km\left(m+1\right)/2$
block diagonal matrix with $j\text{th}$ block
\[
\Sigma_{j}=4\mathbb{E}_{P_{S}}\left[\mathbf{1}\left\{ \mathcal{V}\left(S\right)\in M_{j}\right\} \left(\mathcal{V}\left(S\right)-\underline{A}_{j}^{k}\right)\left(\mathcal{V}\left(S\right)-\underline{A}_{j}^{k}\right)^{T}\right]\text{,}
\]
and $M_{j}=\left\{ M\in\mathbb{R}^{m\left(m+1\right)/2}:j=\arg\min_{j\in\left[k\right]}\left\Vert M-\underline{A}_{l}^{k}\right\Vert \right\} $.
\end{prop}
Under general assumptions regarding $P_{S}$, Proposition 1 provides
the almost sure convergence between $\mathcal{D}_{n}^{k}\left(X^{k}\right)$
and $m_{k}\left(P_{S}\right)$. However, a compactness assumption
on the sample space of $S_{1},\dots,S_{n}$ allows for the quantification
of rates, via the application of Theorems 4--6 of \cite{[13]} to establish
the following result.

\begin{prop}
In addition to the conditions of Proposition 1, assume that $P_{S}$
is compactly supported on $\left\{ S\in\mathbb{S}_{m}:\left\Vert \mathcal{V}\left(S\right)\right\Vert ^{2}\le r\right\} $
for some $r>0$. Then:
\[
\mathbb{E}_{P_{S}}\mathcal{Q}\left(X_{n}^{k},P_{S}\right)-m_{k}\left(P_{S}\right)\le C_{1}n^{-1/2}\text{,}
\]
\[
m_{k}\left(P_{S}\right)-\mathbb{E}_{P_{S}}\mathcal{D}_{n}^{k}\left(X_{n}^{k}\right)\le C_{2}n^{-1/2}\text{, and}
\]
\[
\mathbb{E}_{P_{S}}\mathcal{Q}\left(X_{n}^{k},P_{S}\right)-m_{k}\left(P_{S}\right)=O\left(n^{-1/2}\log^{1/2}n\right)\text{,}
\]
almost surely, where $C_{1}$ and $C_{2}$ are constants that only
depend on $m$, $k$, and $r$.
\end{prop}

Thus far, we have assumed that $k$ is known. However, for unknown
$k$, we require a procedure that estimates its value. Let $k\in\mathbb{K}\subset\mathbb{N}$
and define,
\[
k^{*}=\underset{k\in\mathbb{K}}{\min}\:m_{j}\left(P\right)
\]
for some $P$. We can estimate $k^{*}$ using the Bayesian information
criterion (BIC) inspired estimator: 
\begin{equation} \label{eq:BIC}
k_{n}^{*}=\min_{k\in\mathbb{K}}\:\mathcal{D}_{n}^{k}\left(X_{n}^{k}\right)+m\left(m+1\right)\frac{k\log n}{n}\text{.}
\end{equation}
Via Theorem 8.1 and Corollary 8.2 of \cite{baudry2015estimation}, we have the following
result.
\begin{prop}
Assume the conditions of Proposition 3. If $\mathbb{H}\left[\underline{\mathcal{Q}}\left(\underline{A}^{k},P_{S}\right)\right]$
is positive define for each $k\in\mathbb{K}\subset\mathbb{N}$, then
$\lim_{n\rightarrow\infty}\Pr\left(k_{n}^{*}\ne k^{*}\right)=0$.
\end{prop}

%--------------------------------------------------------------- sec:application
\section{Application} \label{sec:application}

In flat and arid regions of Australia, the high evaporation rates often imply the absence of surface water storage available for human consumption, irrigation, or mining. In most cases, it results in an increased dependence on groundwater, in over-extraction of groundwater resources, and in groundwater level decrease. Depending on the aquifer conditions, it decreases groundwater discharge into streams and limits the availability of shallow groundwater resources for the Groundwater-Dependant Ecosystems/Vegetation (GDE/V) \cite{richardson2011australian}. While it is crucial to monitor GDV health where groundwater resources are exploited, large-scale mapping techniques \cite{doody2017continental} are not multi-temporal, mostly because clouds limit the coverage of the input imagery products (multispectral) during the wetter months. Cloud-insensitive Synthetic Aperture Radar (SAR) data offer an opportunity for monitoring GDEs \cite{castellazzi2019towards} but further research is required to better extract the GDV information it contains.

From SAR data, both intensity and coherence products potentially contain information about GDV:  The like-polarised band VV, dominated by double-bounce and soil-interaction scattering mechanisms (i); the cross-polarised band VH, representing mostly the volumetric scattering and other angular-shifts during signal bounce (ii); and the InSAR coherence matrix CC, derived by comparing the phase of two like-polarised bands (VV) of two subsequent acquisitions (‘repeat path’) (iii). 

GDVs are expected to have a stable canopy over time as compared to non-GDVs, due to their ability to supplement their water requirements using groundwater during times of water deficit and drought. As such, the proportions of volumetric, soil, and double-bounce scattering mechanisms are expected to be relatively stable in time. In \cite{castellazzi2019towards}, an effort is made to leverage this behaviour for classification of GDVs from SAR images, on a pixel-by-pixel basis, where the $\text{SARGDE}_{v1}$ index of a pixel is introduced as 
\begin{equation}
\text{SARGDE}_{v1} = 1/(\sigma_{cc}\sigma_{vh}\mu_{cc}).
\end{equation}
Here, $\sigma_{cc}$ and $\mu_{cc}$ are the standard deviation and mean of InSAR coherence CC for the given pixel, sampled over time, and $\sigma_{vh}$ is the corresponding standard deviation in the linearly-projected VH band intensity values. 
A hypothesis in \cite{castellazzi2019towards} is that a threshold $T$ exists such that GDV locations correspond to pixels with $\text{SARGDE}_{v1} > T$. To examine this hypothesis, a ground truth of GDV locations is provided in the Bureau of Meteorology GDE atlas \cite{GDVatlas}.

In the present study, we step back from direct GDV classification, and focus on exploratory data mining, to better understand the classification task and $\text{SARGDE}_{v1}$ behaviour. 

%---------------------------------------------------------- subsec:methods
\subsection{Methodology} \label{subsec:methods}
Each pixel is represented by a multivariate time-series (of 30 observations in each of the VV and VH bands, and 29 observations in the CC product), acquired over one year, in the Mount Gambier region of Australia, in 2017. At the $\approx 30$m resolution, the image dimensions are $2044\times1433$, so that the number of time-series (pixels) in the sample is $n = 2{,}929{,}052.$ We describe the data products in detail in Section \ref{subsec:preprocessing}. From these products, the sample finite \hbox{$\ell$-lag} autocovariance matrices are computed and transformed to the log-Cholesky space via \eqref{eq:log-chol-vec}. Prior to this transformation, downsampling is performed via local averaging within $p \times p$ patches, to decrease variability. The patch size, $p$, and lag length, $\ell$, are hyperparameters that we choose by maximising the adjusted Rand index \cite{hubert1985comparing} for the agreement between $k$-means clusters and ground truth locations.

Given the obtained hyperparameters, an optimal number of clusters $k^*_n$ are chosen via the BIC inspired estimator \eqref{eq:BIC}. We then seek to further justify this choice, by estimating the $\text{SARGDE}_{v1}$ variability explained by the $k^*_n$ clusters, via $k^*_n$-way ANOVA. We compare this to a 2-way ANOVA with simplifying restriction $k = 2$, to produce an estimate of the additional variability explained by the $k^*_n > 2$ clusters.

Finally, we visualise, in a log-Cholesky space, the pixels that belong to $k$-means clusters that have more than 5\% empirical probability of overlap with GDV. In the same space, we visualise the sample quartiles of $\text{SARGDE}_{v1}$. From this, we draw conclusions about $\text{SARGDE}_{v1}$, and make suggestions for future efforts to classify GDV.  

For repeatability, all scripts (in the R programming language) and pre-processed data are available in an online repository at \cite{kmspd}.

%---------------------------------------------------------- subsec:preprocessing
\subsection{Data pre-processing} \label{subsec:preprocessing}
A total of 30 temporally consecutive Sentinel-1A Interferometric Wide (IW) images in Single-Look-Complex (SLC) format were downloaded via the Alaskan Satellite Facility (ASF) web portal \cite{ASFWeb} and processed similarly to \cite{castellazzi2019towards}. The time-series consist of images acquired along the same orbital track and Line-Of-Sight (LOS) angle, which facilitates the interpretation of SAR intensity change over time (i.e., no significant LOS change to take into account or compensate for) and allows the creation of Interferometric SAR (InSAR) coherence maps \cite{zebker1992decorrelation}. The 12-day repeat path of Sentinel satellites guarantees $\approx30$ intensity images per year, and $\approx29$ coherence maps per year.

Sentinel-1 IW images were processed using SARscape 5.5 \cite{SARscape}. The images are multi-looked (i.e., decrease in resolution) with a factor $8/2$ in Range/Azimuth to produce a regular matrix grid at $\approx30$m resolution. This reduces data size and granular noise (‘speckle’) inherent to SAR data. All images were co-registered and spatio-temporally filtered \cite{de1997radar} to remove residual noise. Images are then calibrated and converted into backscatter coefficients following a Gamma Nought calibration (correction for local incidence angle variations using the ALOS-3D Digital Elevation Model \cite{takaku2014generation}), and projected along a linear scale. Coherence matrices are computed at the same resolution as the intensity matrices ($30$m) and with a $5\times5$ pixel analysis window. They were produced in a ‘time-line’ process, where each coherence matrix is produced by matching with the subsequent image.

%------------------------------------------------------------------- sec:results
\subsection{Results} \label{subsec:results}

\begin{figure}
    \centering
    \includegraphics[scale = 0.6]{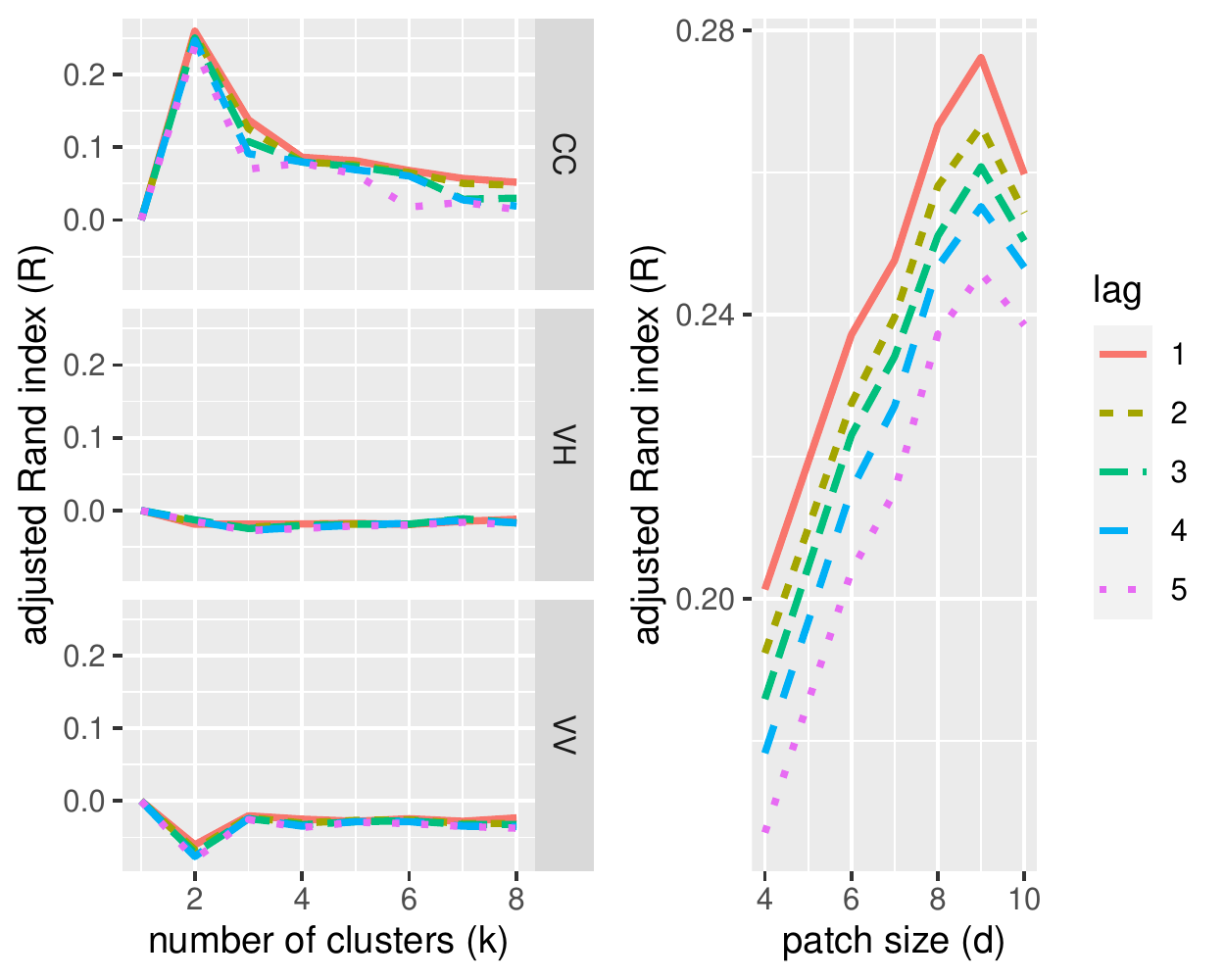}
    \caption{The left panel gives the adjusted Rand index for lags $\ell \in [5]$, clusters $k \in [8]$ and each of the products CC, VH and VV. The right panel gives that for $k=2$, in the CC product, and patch sizes $p \in \{4,5,\ldots,10\}$.}
    \label{fig:rand}
\end{figure}

As shown in Figure \ref{fig:rand}, the hyperparameters maximising the adjusted Rand index $R$, for agreement of $k$-means classes with GDV locations, were $\ell = 1$ and $p = 9$. Note that the spike in $R$, seen in Figure \ref{fig:rand} near $k=2$, occured only in the CC product, while VV and VH showed very low overlap with GDV. We expect, a priori, that the CC product carries the majority of accessible information about vegetation \cite{castellazzi2019towards}, and this result appears to confirm this expectation. Furthermore, it is feasible that the observed decrease in $R$, for $k > 2$, in the CC product, is due to GDV qualities being split between multiple $k$-means classes. From this point on, we focus our analysis on the CC product.

The BIC inspired estimator \eqref{eq:BIC}, with the obtained hyperparameters, gave $k^*_n=15$ when explored over $k \in [50]$. For the corresponding $15$-way ANOVA, the adjusted coefficient of multiple correlation was $0.532$, indicating that $\approx53\%$ of the variability in $\text{SARGDE}_{v1}$ is explained by the $15$ clusters. For comparison, only $\approx20\%$ of the variability is explained when $k=2$ (that is, when fitting $k$-means with only $2$ clusters). This $33\%$ increase in explained $\text{SARGDE}_{v1}$ variability further justifies the use of $k > 2$, at least for the purpose of understanding $\text{SARGDE}_{v1}$.

\begin{figure}
    \centering
    \includegraphics[scale = 0.25]{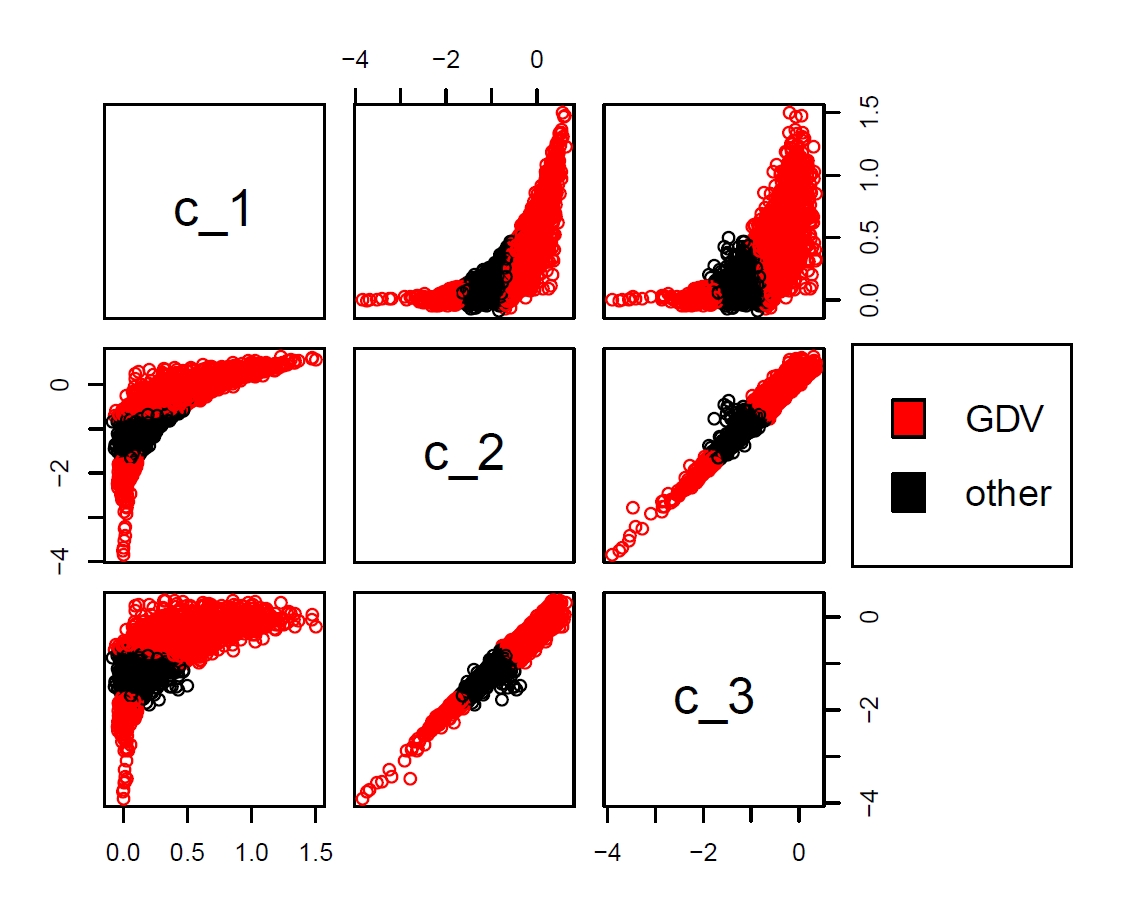}
    \caption{Cross-sectional projection scatter plots of the $n$ time-series (pixels), represented in the 3-dimensional space of autocovariance matrices transformed via \eqref{eq:log-chol-vec}. Pixels in $k$-means clusters $1,2,5$ and $12$ (clusters with greater than 5\% empirical probability of GDV) are coloured black, and all others are red.}
    \label{fig:overlap-gdv}
\end{figure}

Of the 15 $k$-means classes, only four have more than $5\%$ of pixels overlapping with GDV. These are clusters $1,2,5$ and $12$. In Figure \ref{fig:overlap-gdv}, these four clusters are coloured black, and the remaining are coloured red. Time-series (pixels) are represented in Figure \ref{fig:overlap-gdv} by the transformation \eqref{eq:log-chol-vec} of their autocovariance matrices. Notice that the 4 classes are neighbours, and occupy a band near the middle of the larger cluster. Pixels outside of this band have a low (less than $5\%$) proportion of overlap with GDV. 

\begin{figure}
    \centering
    \includegraphics[scale = 0.25]{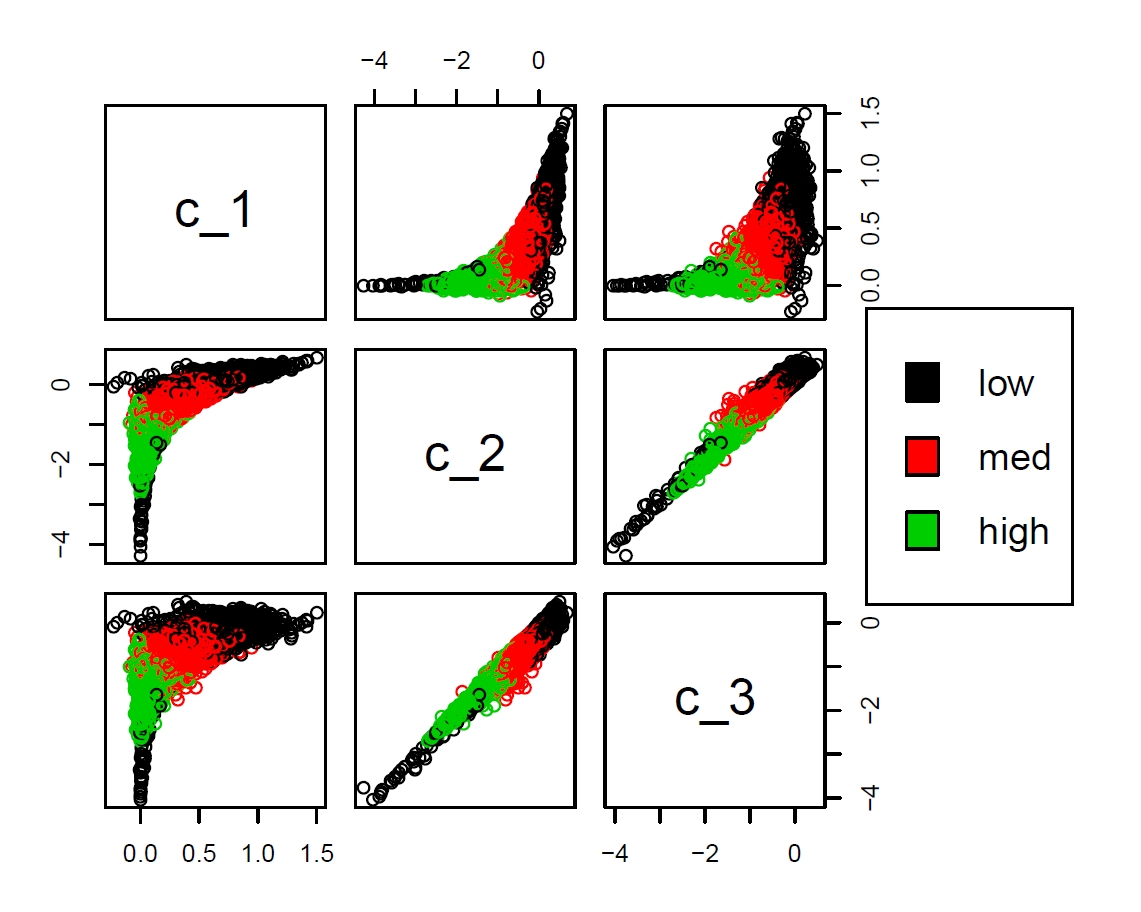}
    \caption{Cross-sectional projection scatter plots of the $n$ time-series (pixels), represented in the 3-dimensional space of autocovariance matrices transformed via \eqref{eq:log-chol-vec}. Pixels whose $\text{SARGDE}_{v1}$ is in the lower 25\% sample quartile are coloured black, the middle 50\% red, and the upper 25\% green.}
    \label{fig:sargde-qnt}
\end{figure}

For comparison with the $\text{SARGDE}_{v1}$ quartiles, Figure \ref{fig:sargde-qnt} colours the lower 25\% quartile black, the middle 50\% red, and the upper 25\% green. We see that high $\text{SARGDE}_{v1}$ pixels occupy a band near the middle of the larger cluster (green in Figure \ref{fig:sargde-qnt}), but that this band is wide enough that it appears to transgress into the regions with less than 5\% empirical probability of GDV (red, in Figure \ref{fig:overlap-gdv}).

\begin{figure}
    \centering
    \includegraphics[scale=0.5]{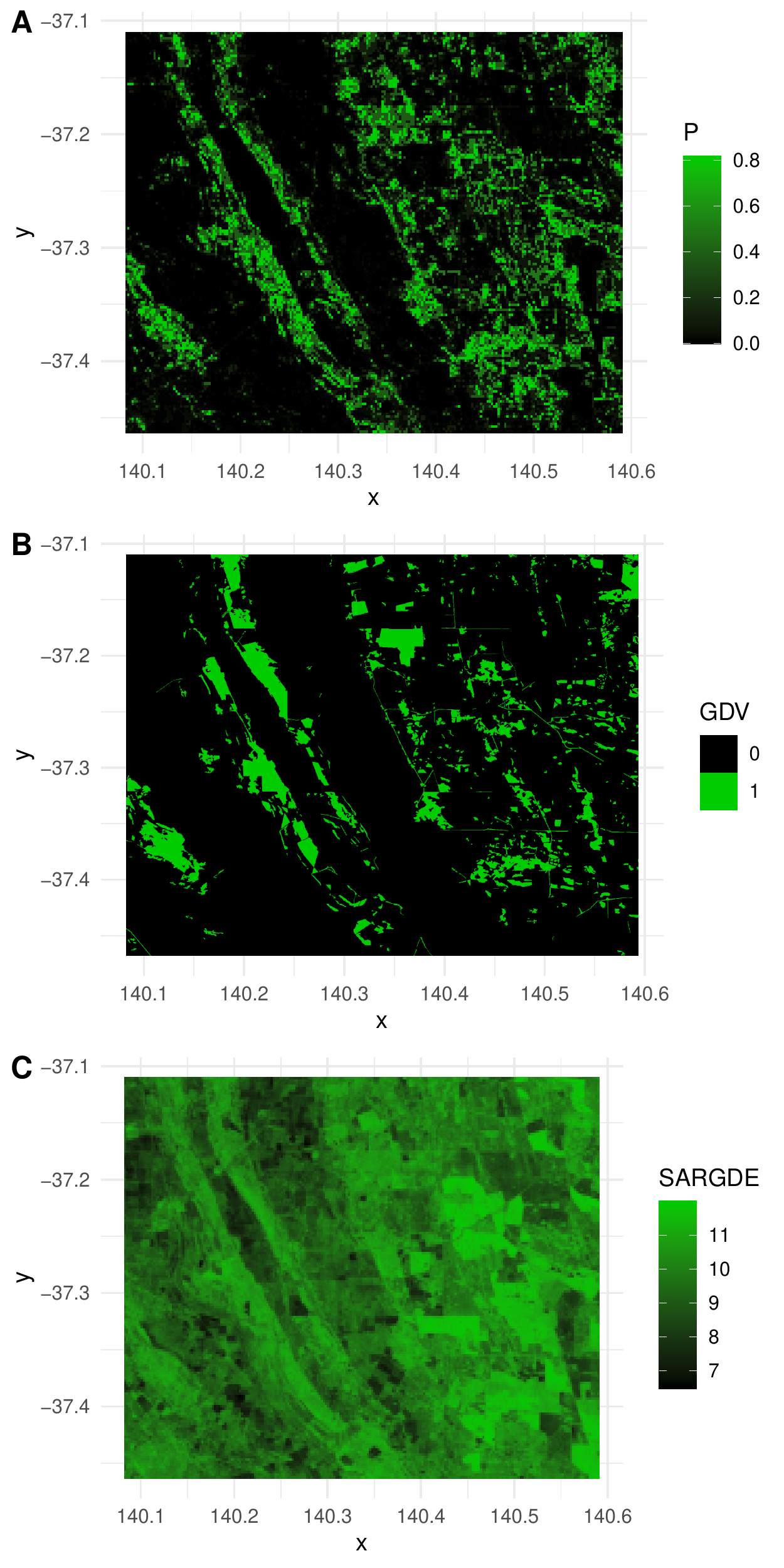}
    \caption{(A) The 15 $k$-means classes, coloured by empirical probability $P$ of overlap with GDV; (B) the Bureau of Meteorology GDV atlas ground truth labels; (C) the $\text{SARGDE}_{v1}$ index.}
    \label{fig:three_maps}
\end{figure}

%---------------------------------------------------------------- sec:discussion
\section{Discussion} \label{sec:discussion}

 The observation that the highest $\text{SARGDE}_{v1}$ index values transgress into regions of low empirical probability of GDV, suggests a potentially better approach than using a single threshold, $T$, above which $\text{SARGDE}_{v1}$ classifies pixels as GDV. Instead, a lower threshold $T_{\ell}$ and upper threshold $T_u$, may be sought, providing an optimal $\text{SARGDE}_{v1}$ interval, in terms of GDV classification performance. Visually, we observe in Figure \ref{fig:three_maps}, that higher values of $\text{SARGDE}_{v1}$ do not necessarily correspond more to GDV. This is our suggestion for future efforts to improve on $\text{SARGDE}_{v1}$. Alternatively, a classifier can be constructed from the $k$-means clusters, directly, instead of using the raw $\text{SARGDE}_{v1}$ values.

\subsection{Future work}

\begin{itemize}
    \item We have treated the autocovariance matrices of time-series (pixels) as stationary, though more information may be obtained by capturing dynamics via covariance trajectories in the space of SPD matrices. See, e.g., \cite{zhang2018rate}.
    \item Due to the one-to-one mapping of any SPD matrix to a real vector, via a differentiable transformation, one can endow the space of SPD matrices with any distribution on multivariate real numbers, such as the Gaussian distribution, and obtain a distribution over the SPD vector via a transformation of variables construction. This then allows for the conduct of model-based clustering, via the methods of \cite{mclachlan2004finite}.
    \item A similar construction to \eqref{eq:log-chol-vec} exists for the log-Euclidean metric norm, in which the off diagonals are mapped to twice their value. Thus, we expect similar properties to those proved in Section \ref{eq:log-chol-vec} to hold.
    \item This approach to clustering has not made use of information regarding the spatial dependence between pixels across patches. A variety of methods exist that may make use of this information, such as, for example, spatial smoothing via Markov random fields. 
\end{itemize}

%---------------------------------------------------------------- Acknowledgment
%\section*{Acknowledgment}

%\section*{References}
\bibliographystyle{./IEEEtran}
\bibliography{bibliography}

%\vspace{12pt}
%\color{red}
%IEEE conference templates contain guidance text for composing and %formatting conference papers. Please ensure that all template text is %removed from your conference paper prior to submission to the conference. %Failure to remove the template text from your paper may result in your %paper not being published.

\end{document}